\documentclass[letterpaper]{article} 
\usepackage{aaai25}  

\usepackage{times}  
\usepackage{helvet}  
\usepackage{courier}  
\usepackage[hyphens]{url}  
\usepackage{graphicx} 
\usepackage{natbib}  
\usepackage{caption} 
\usepackage{algorithm}
\usepackage{algorithmic}

\usepackage{booktabs}
\usepackage{siunitx}
\usepackage{array}
\usepackage{subcaption}
\usepackage{multirow}
\usepackage{amsmath}
\usepackage{amssymb}

\usepackage[percent]{overpic}  

\usepackage{newfloat}
\usepackage{listings}
\DeclareCaptionStyle{ruled}{labelfont=normalfont,labelsep=colon,strut=off} 
\floatstyle{ruled}
\newfloat{listing}{tb}{lst}{}
\floatname{listing}{Listing}
\title{SmokeNet: Efficient Smoke Segmentation Leveraging Multiscale Convolutions and Multiview Attention Mechanisms}
\author {
    Xuesong Liu\textsuperscript{\rm 1},
    Emmett J. Ientilucci\textsuperscript{\rm 1}
}
\affiliations {
    \textsuperscript{\rm 1}Rochester Institute of Technology, Rochester, New York, USA\\
    {\{xl2088, ejipci\}@rit.edu}
}

\usepackage{bibentry}

\begin{document}

\maketitle

\begin{abstract}
Efficient segmentation of smoke plumes is crucial for environmental monitoring and industrial safety, enabling the detection and mitigation of harmful emissions from activities like quarry blasts and wildfires. Accurate segmentation facilitates environmental impact assessments, timely interventions, and compliance with safety standards. However, existing models often face high computational demands and limited adaptability to diverse smoke appearances, restricting their deployment in resource-constrained environments. To address these issues, we introduce SmokeNet, a novel deep learning architecture that leverages multiscale convolutions and multiview linear attention mechanisms combined with layer-specific loss functions to handle the complex dynamics of diverse smoke plumes, ensuring efficient and accurate segmentation across varied environments. Additionally, we evaluate SmokeNet's performance and versatility using four datasets, including our quarry blast smoke dataset made available to the community. The results demonstrate that SmokeNet maintains a favorable balance between computational efficiency and segmentation accuracy, making it suitable for deployment in environmental monitoring and safety management systems. By contributing a new dataset and offering an efficient segmentation model, SmokeNet advances smoke segmentation capabilities in diverse and challenging environments.

\end{abstract}

\section{Introduction}

Accurate smoke segmentation is essential for environmental monitoring and industrial safety, facilitating early fire detection, pollution control, and the assessment of emissions from various industrial activities. Among these, smoke resulting from quarry blasting presents unique and complex challenges due to its variable characteristics, including irregular shapes, mixed textures of dust and debris, and varying levels of opacity. These complexities necessitate a segmentation model capable of distinguishing smoke from surrounding elements and accurately capturing its dynamic structure and boundaries across diverse industrial settings. Addressing these challenges requires advanced deep learning architectures that can effectively handle the intricate and dynamic nature of smoke plumes, ensuring precise and reliable segmentation in real-world applications.

Semantic object segmentation has significantly advanced with deep learning architectures like UNet \cite{ronneberger2015u} and Fully Convolutional Networks (FCNs) \cite{long2015fully}, which utilize encoder-decoder structures to capture and reconstruct complex image features. However, these models often struggle with dynamic scenes such as smoke plumes, influenced by factors like wind, humidity, and varying smoke sources. This variability limits their adaptability for transient, high-variability conditions essential for low-latency monitoring. To address this, efficient CNN architectures like MobileNet \cite{howard2017mobilenets}, EfficientNet \cite{tan2019efficientnet}, and ShuffleNet \cite{zhang2018shufflenet} have been developed to achieve performance through reduced computational complexity. Additionally, Vision Transformers (ViTs) \cite{dosovitskiy2020image} enhance the ability to capture global and contextual information, though they typically require higher computational resources. While these architectures offer significant improvements, they often involve trade-offs between computational efficiency and segmentation accuracy, particularly in resource-constrained environments where continuous, practical smoke segmentation is necessary.

Specialized studies in smoke segmentation, such as the deep Smoke Segmentation (DSS) model~\cite{yuan2019deep}, Frizzi et al.~\cite{frizzi2021convolutional}, and Yuan et al.~\cite{yuan2023lightweight}, have explored advanced techniques to address these challenges. The DSS model employs a two-path fully convolutional network to extract global context, enhancing segmentation accuracy but increasing computational complexity. Frizzi et al. introduced a VGG16-based model that attempts to improve smoke plume segmentation performance compared to image processing techniques, yet this comes with higher model parameters and complexity. Yuan et al.~\cite{yuan2023lightweight} proposed a lightweight model incorporating attention mechanisms to replace the global context extraction from two-path FCNs, achieving significant parameter reduction while maintaining strong performance. As far as we know, Yuan's lightweight paper is the most recent work with a lightweight model and good performance targeting smoke segmentation, having already significantly reduced the parameters compared to previous smoke segmentation networks. To further improve smoke segmentation efficiency with a low-parameter model, we introduce SmokeNet, a novel and efficient UNet-based architecture specifically designed to meet the unique demands of smoke segmentation in both synthetic and real-world environments, with a particular focus on quarry smoke. Our contributions include:

\begin{itemize}
    \item \textbf{Multiscale Convolutions with Rectangular Kernels:} SmokeNet integrates a multiscale convolution module using rectangular-shaped kernels alongside standard kernels, allowing it to adapt to the irregular shapes often seen in smoke. This approach provides better spatial information, with vertically oriented kernels capturing the tall, narrow shapes typical of wildfire smoke and horizontally oriented kernels suited for the wide, low plumes found in quarry blast smoke.
    \item \textbf{Lightweight Multiview Linear Attention:} To enhance feature integration without imposing high computational costs, SmokeNet incorporates a linear attention mechanism with multi-view element-wise multiplication, enabling the model to selectively attend to both spatial and channel-wise features. This design preserves accuracy while significantly reducing the parameter count, allowing smoke segmentation even in GPU-constrained settings.
    \item \textbf{Layer-Specific Loss:} To optimize feature refinement, we introduce a layer-specific loss strategy that minimizes feature gaps across the network’s layers, fostering more detailed and precise feature learning. This approach enhances segmentation accuracy by aligning intermediate feature representations throughout the network, thereby supporting consistent and refined feature extraction without increasing model complexity.
\end{itemize}


\section{Related Work}

\subsection{Encoder-Decoder Architectures}

The encoder-decoder paradigm has been pivotal in advancing image segmentation tasks. UNet++ \cite{zhou2018unet++} enhanced this framework by introducing nested and dense skip pathways, effectively bridging semantic gaps between encoder and decoder features and improving segmentation accuracy through deep supervision. This indicates an improved feature flow among the different stages of the model compared to the original UNet \cite{ronneberger2015u}. Additionally, UNet++ incorporates a pruned decoder to reduce the number of parameters, enhancing computational efficiency.

Expanding upon traditional encoder-decoder frameworks, ERFNet \cite{romera2017erfnet} is designed to deliver high accuracy with reduced computational complexity. ERFNet utilizes factorized convolutions and residual connections to streamline the network, making it suitable for applications such as autonomous driving and robotics where real-time processing is essential. Similarly, DFANet \cite{li2019dfanet} introduces a dual attention mechanism that captures both spatial and channel-wise dependencies, enhancing feature representation and improving performance in semantic segmentation and object detection tasks. DFANet achieves a balance between speed and segmentation performance by aggregating discriminative features through a lightweight backbone and multi-scale feature propagation.

Inspired by enhanced feature learning, Deep Smoke Segmentation (DSS) \cite{yuan2019deep} employs a dual-path encoder-decoder structure based on fully convolutional networks, specifically designed for smoke segmentation. This architecture achieves good performance; however, the parameter count remains relatively high. Similarly, Frizzi et al. \cite{frizzi2021convolutional} developed a convolutional neural network using VGG architectures with multiple kernel sizes to capture both global context and fine spatial details. This approach enhances segmentation performance across diverse datasets by effectively handling the dynamic and amorphous nature of smoke plumes, though it results in significant parameter counts.

To comprehensively extract global context features, attention-based mechanisms have been integrated into encoder-decoder architectures. Attention UNet \cite{oktay2018attention} incorporates attention gates into the UNet architecture, enabling the model to focus on relevant target structures and thereby improve segmentation accuracy. This selective focus helps in better delineating smoke regions from complex backgrounds. Additionally, CGNet \cite{wu2020cgnet} introduces attention modules within a Context Guided Network framework, prioritizing salient features for efficient semantic segmentation on mobile devices. These attention mechanisms enhance the model's ability to discern important features, thereby improving overall segmentation performance.

More recently, MobileViTv2 \cite{mehta2022separable} introduces a hybrid approach that combines convolutions with Vision Transformers to enhance feature representation for semantic segmentation tasks. By addressing the latency issues commonly associated with multi-headed self-attention (MHA) mechanisms, MobileViTv2 employs a separable self-attention mechanism with linear complexity. This improvement makes the model more practical for resource-constrained environments while maintaining competitive segmentation performance.



\subsection{Lightweight Segmentation Models}

Segmentation in GPU-constrained environments necessitates lightweight models that balance accuracy with computational efficiency. Several architectures have been developed to optimize computational resources through innovative techniques. MobileNet \cite{howard2017mobilenets}, for instance, employs depthwise separable convolutions to reduce the number of parameters and computational load, making it suitable for mobile and embedded applications. Similarly, ShuffleNet \cite{zhang2018shufflenet} introduces pointwise group convolutions and channel shuffle operations to achieve high efficiency without significant accuracy loss. EfficientNet \cite{tan2019efficientnet} utilizes a compound scaling method that uniformly scales network depth, width, and resolution, providing a family of models that offer a balance between performance and efficiency.

Building upon these lightweight foundations, UNeXt-S \cite{valanarasu2022unext} is another lightweight model that has shown promise in medical image segmentation by incorporating efficient convolutional operations and attention mechanisms. Although these models were not initially designed specifically for smoke segmentation, their emphasis on computational efficiency makes them suitable candidates for efficient applications in this domain.

MALUNet \cite{ruan2022malunet} and LEDNet \cite{wang2019lednet} exemplify the advancement of lightweight segmentation models tailored for specific tasks. MALUNet introduces a lightweight architecture designed for skin lesion segmentation by integrating four specialized modules: Dilated Gated Attention (DGA), External Attention (IEA), Channel Attention Block (CAB), and Spatial Attention Block (SAB). These modules collectively enable the network to efficiently extract and fuse both global and local features while significantly reducing the number of parameters and computational complexity. By adopting a U-shape architecture, MALUNet achieves competitive segmentation performance with minimal resource requirements, making it suitable for deployment in resource-constrained clinical environments.

LEDNet further refines lightweight segmentation by employing an asymmetric encoder-decoder architecture. It utilizes channel split and shuffle operations within residual blocks to minimize computational costs. Additionally, an Attention Pyramid Network (APN) is integrated into the decoder to enhance segmentation accuracy without increasing model complexity. This design ensures that LEDNet remains highly efficient, making it ideal for real-time applications on mobile devices where computational resources are limited.

Recent advancements have specifically targeted the smoke segmentation task with a focus on enhancing parameter efficiency. Yuan et al. \cite{yuan2023lightweight} introduced a refined lightweight model optimized for efficient smoke segmentation. This model integrates attention mechanisms to efficiently extract salient features while minimizing computational demands, effectively leveraging the strengths of both lightweight architectures and attention mechanisms to improve performance on smoke-related tasks.

Despite advances in efficient models, evaluating their performance is limited by the scarcity of diverse smoke segmentation datasets. Many studies report good results on synthetic data or specific scenarios like fire-smoke segmentation, but the limited diversity and size of current datasets impede comprehensive assessment of model generalizability. Expanding and diversifying smoke-related datasets to include complex environments such as quarry smoke is essential. This will enable the development and validation of models that can generalize across varied environments, enhancing the robustness and applicability of smoke segmentation techniques in practical applications.

\begin{figure*}[htbp]
    \centering
    \begin{minipage}{0.45\linewidth}
        \centering
        \begin{subfigure}[b]{\linewidth}
            \centering
            \includegraphics[width=\linewidth]{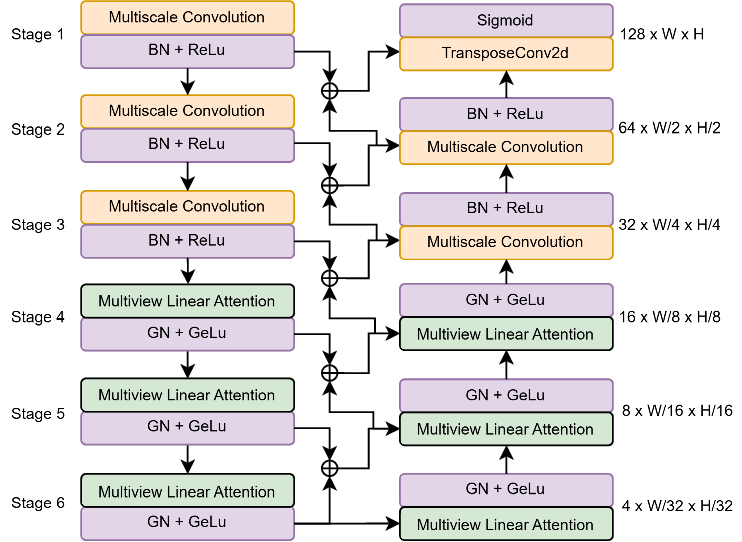}
            \caption{Model Architecture}
            \label{fig:arch_overview}
        \end{subfigure}
    \end{minipage}
    \hfill
    \begin{minipage}{0.50\linewidth}
        \centering
        \begin{subfigure}[b]{\linewidth}
            \centering
            \includegraphics[width=\linewidth]{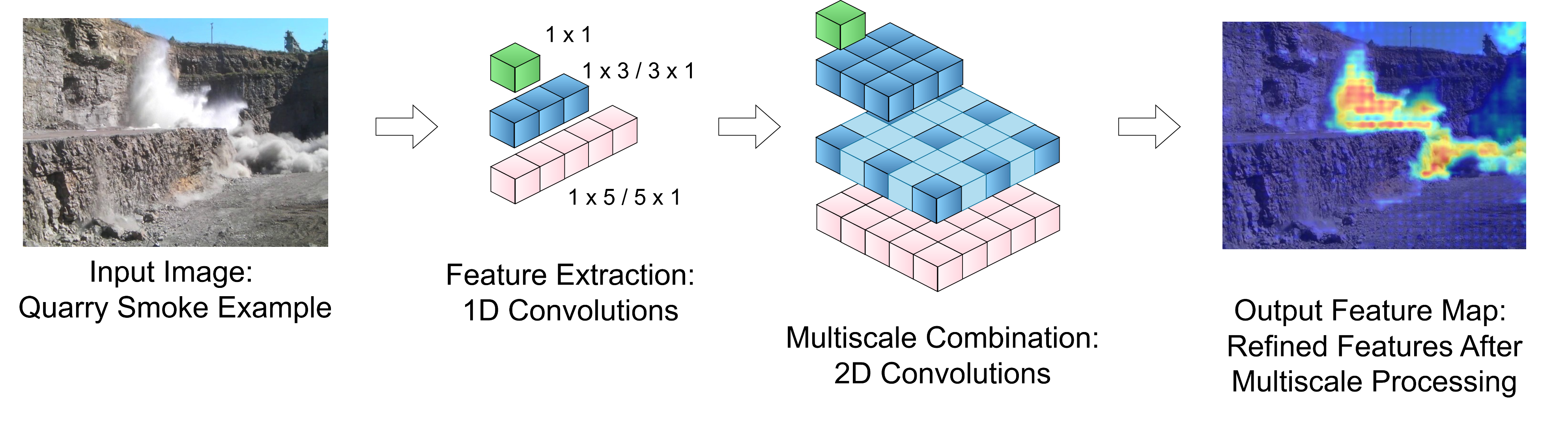}
            \caption{Multiscale Module}
            \label{fig:multiscale_module}
        \end{subfigure}
        
        \vspace{0.5cm} 
        
        \begin{subfigure}[b]{\linewidth}
            \centering
            \includegraphics[width=\linewidth]{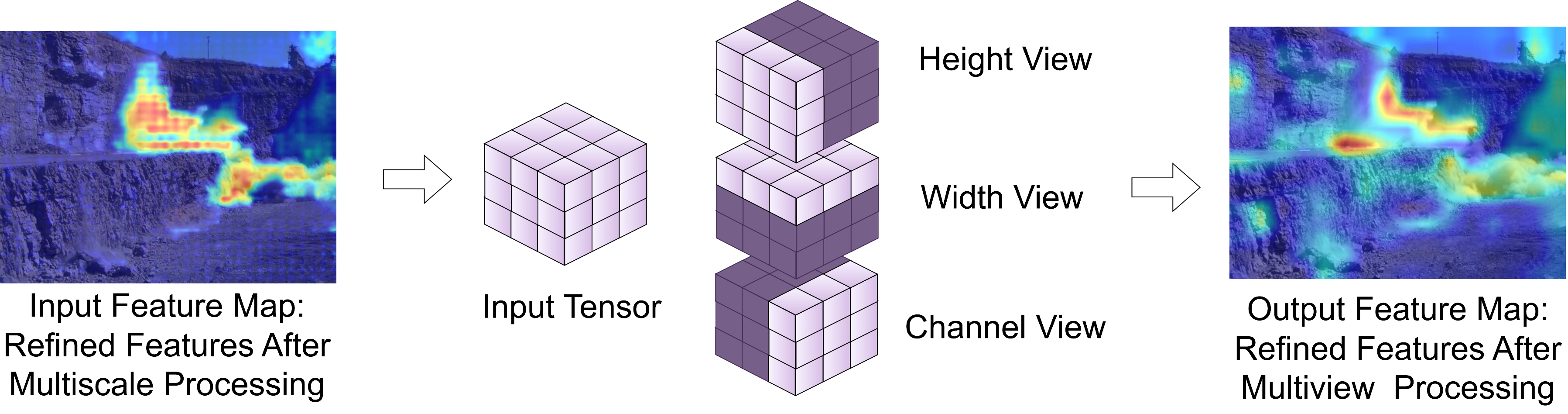}
            \caption{Multiview Module}
            \label{fig:multiview_module}
        \end{subfigure}
    \end{minipage}
    
    \caption{Overview of SmokeNet's Architecture. (a) The overall model architecture integrating multiscale convolutions and multiview attention mechanisms.  (b) The Multiscale Module capturing spatial information at various scales for accurate smoke segmentation.  (c) The Multiview Module enhancing feature refinement through attention mechanisms.}
    \label{fig:combined_architecture}
\end{figure*}
\section{Methodology}

SmokeNet is an architecture developed for smoke segmentation in complex environments, particularly addressing the dynamic characteristics of quarry smoke. Inspired by UNet, the network employs an encoder-decoder structure organized into six stages. The encoder (Stages 1-6) focuses on extracting hierarchical features through a combination of multiscale convolutions and attention mechanisms. The decoder (Stages 1-6) progressively reconstructs the segmentation map while incorporating multiview linear attention mechanisms and skip connections to integrate features from corresponding encoder stages. This design ensures SmokeNet effectively handles the variability and structural complexities of smoke plumes (Figure~\ref{fig:arch_overview}).

\subsection{Encoder}

\paragraph{Multiscale Feature Extraction}

The encoder stages (Stages 1-3) are responsible for extracting features across multiple spatial scales, as shown in Figures~\ref{fig:arch_overview} and~\ref{fig:multiscale_module}, enabling the model to effectively capture the variability inherent in smoke patterns. This multiscale extraction is facilitated by a dedicated multiscale module, which processes input feature tensors through a series of convolutional operations with varying kernel sizes, followed by batch normalization and activation functions.

The Multiscale Module employs 1D convolutional layers with diverse kernel sizes, including \(1 \times 1\), \(1 \times 3\), \(3 \times 1\), \(1 \times 5\), and \(5 \times 1\). These convolutions are applied sequentially to the input tensor \(\mathbf{F}\), resulting in multiple feature maps that capture different spatial extents and orientations of smoke plumes. For instance, the \(1 \times 3\) and \(3 \times 1\) convolutions are adept at capturing elongated features in horizontal and vertical directions, respectively, while the \(1 \times 5\) and \(5 \times 1\) convolutions capture broader spatial contexts.

To construct larger and more complex 2D convolution operators efficiently, sequential 1D convolutions are employed. The equivalent 2D convolution operators, such as \(3 \times 3\), \(3 \times 5\), \(5 \times 3\), and \(5 \times 5\), are defined by applying two 1D convolutions in orthogonal directions:
\begin{align*}
\text{Conv}_{3\times3} &= \text{Conv}_{3\times1} \circ \text{Conv}_{1\times3}, \\
\text{Conv}_{3\times5} &= \text{Conv}_{3\times1} \circ \text{Conv}_{1\times5}, \\
\text{Conv}_{5\times3} &= \text{Conv}_{5\times1} \circ \text{Conv}_{1\times3}, \\
\text{Conv}_{5\times5} &= \text{Conv}_{5\times1} \circ \text{Conv}_{1\times5}.
\end{align*}
Here, each operator is constructed by first applying a 1D convolution along one axis, followed by another 1D convolution along the orthogonal axis. For instance, \(\text{Conv}_{3\times3}\) is achieved by applying \(\text{Conv}_{1\times3}\) (horizontal) followed by \(\text{Conv}_{3\times1}\) (vertical).

These operations, especially the rectangular kernel sizes targeting horizontal or vertical features, enable the module to capture a wide range of smoke shapes, from narrow and tall plumes common in campfires to wide and elongated patterns resulting from quarry blasts.

Let us denote the input tensor to the encoder as \(\mathbf{F} \in \mathbb{R}^{N \times C \times H \times W}\), where \(N\) represents the batch size, \(C\) is the number of channels, and \(H\) and \(W\) denote the spatial height and width of the feature maps, respectively.

For each stage, including the Multiscale Module, the input—either from the original image (Stage 1) or the output of the previous stage—is split into four chunks along the channel dimension. Each chunk \(\mathbf{F}_i \in \mathbb{R}^{N \times \frac{C}{4} \times H \times W}\) is processed independently using specific operations, as follows:
\begin{align*}
\mathbf{F}_1' &= \text{RELU}\left( \mathbf{F}_{1} \right), \quad \text{(identity mapping)} \\
\mathbf{F}_2' &= \text{RELU}\left( \text{Conv}_{1\times1}\left( \mathbf{F}_{2} \right) \right), \quad \text{($1 \times 1$ convolution)} \\
\mathbf{F}_3' &= \text{RELU}\left( \text{Conv}_{\text{selected}}\left( \mathbf{F}_{3} \right) \right), \quad \text{(non-dilated convolution)} \\
\mathbf{F}_4' &= \text{RELU}\left( \text{Conv}_{\text{selected}}^{\text{dilated}}\left( \mathbf{F}_{4} \right) \right), \quad \text{(dilated convolution)}.
\end{align*}

The convolutional operations in \(\text{Conv}_{\text{selected}}\) include a range of kernel sizes and their sequential combinations to emulate 2D convolutions:
\begin{align*}
\text{Convs} = \Big[ &\text{Conv}_{1\times3},\ \text{Conv}_{3\times1},\ \text{Conv}_{1\times5},\ \text{Conv}_{5\times1}, \\
&\text{Conv}_{3\times3},\ \text{Conv}_{3\times5},\ \text{Conv}_{5\times3},\ \text{Conv}_{5\times5} \Big].
\end{align*}

To address dimensional alignment, all outputs from the selected kernel operations are normalized to consistent dimensions (\(N \times \frac{C}{4} \times H \times W\)) before concatenation. This is achieved by applying a \(1 \times 1\) convolution to adjust the channel count and using appropriate padding or cropping to match the spatial dimensions. These steps ensure compatibility during feature integration, preventing dimensional mismatches and enabling stable multiscale feature fusion, while maintaining efficiency across stages.

After processing, each chunk undergoes batch normalization to stabilize the learning process:
\begin{align*}
\mathbf{F}_i'' = \text{BatchNorm}\left( \mathbf{F}_i' \right), \quad \text{for } i = 1, 2, 3, 4.
\end{align*}

The outputs from all four chunks are concatenated along the channel dimension to form the combined feature map:
\begin{align*}
\mathbf{F}' &= \text{Concat}\left( \mathbf{F}_1'',\ \mathbf{F}_2'',\ \mathbf{F}_3'',\ \mathbf{F}_4'' \right).
\end{align*}

An identity mapping with activation is applied to the concatenated feature map to produce the final feature map:
\begin{align*}
\mathbf{F}_{\text{final}} &= \text{RELU}\left( \text{Proj}\left( \mathbf{F}' \right) \right),
\end{align*}
where \(\text{Proj}(\cdot)\) is a \(1 \times 1\) convolution if the channel dimensions of \(\mathbf{F}\) and \(\mathbf{F}'\) differ, otherwise it is the identity function.

Finally, a \(2 \times 2\) Max Pooling operation reduces the spatial dimensions:
\[
\mathbf{F}_{\text{pooled}} = \text{MaxPool}\left( \mathbf{F}_{\text{final}},\ 2 \times 2 \right).
\]

By integrating batch normalization and RELU activation functions throughout the module, the encoder effectively captures both local and global contextual features while ensuring stable and efficient learning. The use of larger kernels and dilated convolutions in \(\mathbf{F}_4'\) captures broader contextual information, and the inclusion of batch normalization after each path's output normalizes feature distributions, facilitating deeper network training.

As shown in Figure~\ref{fig:multiscale_module}, the encoder of SmokeNet systematically extracts features through a combination of multiscale convolutions, dilated convolutions, activation functions, batch normalization, and strategic feature fusion. The incorporation of these elements enhances the model's ability to capture the variability and structural complexity of smoke plumes in various environments.

\paragraph{Multiview Linear Attention Mechanism}
In encoder stages 4-6, multiview linear attention mechanisms are integrated to refine the encoded features, as shown in Figure~\ref{fig:multiview_module}, focusing on essential elements while maintaining computational efficiency. SmokeNet employs the multiview attention mechanism across different dimensions of the feature tensor—channel, height, and width—providing a comprehensive enhancement of key features.

Let the input tensor at stage \(k\) of the encoder be \(\mathbf{F}^k \in \mathbb{R}^{N \times C \times H \times W}\). For the multiview operation, the tensor is split into four equal chunks along the channel dimension:
\[
\mathbf{F}_{\text{split}, i} = \{\mathbf{F}_1, \mathbf{F}_2, \mathbf{F}_3, \mathbf{F}_4\},
\]
where each chunk \(\mathbf{F}_i \in \mathbb{R}^{N \times \frac{C}{4} \times H \times W}\).

Each chunk undergoes a distinct processing operation involving element-wise multiplication with an attention map computed via softmax activation over specific dimensions:
\begin{align*}
\mathbf{F}_1' &= \mathbf{F}_1 \quad \text{(Identity Mapping)}, \\
\mathbf{F}_2' &= \sigma_{\text{spatial}}(\mathbf{F}_2) \odot \mathbf{F}_2, \\
\mathbf{F}_3' &= \sigma_{\text{height-channel}}(\mathbf{F}_3) \odot \mathbf{F}_3, \\
\mathbf{F}_4' &= \sigma_{\text{width-channel}}(\mathbf{F}_4) \odot \mathbf{F}_4,
\end{align*}
where \(\sigma_{\text{dimension}}(\cdot)\) denotes the softmax activation applied over the specified dimensions, and \(\odot\) represents element-wise multiplication.

The outputs from all four chunks are concatenated along the channel dimension to form the combined feature map:
\begin{align*}
\mathbf{F}' &= \text{Concat}\left( \mathbf{F}_1',\ \mathbf{F}_2',\ \mathbf{F}_3',\ \mathbf{F}_4' \right).
\end{align*}

This concatenated feature map is then processed by a pointwise convolution to integrate the multiscale features into a unified representation:
\begin{align*}
\mathbf{F}_{\text{out}} &= \text{Conv}_{1\times1}\left( \mathbf{F}' \right).
\end{align*}

Following the convolution, layer normalization and GELU activation are applied to \(\mathbf{F}_{\text{out}}\). Finally, a \(2 \times 2\) max pooling operation is performed to reduce the spatial dimensions, resulting in:
\[
\mathbf{F}_{\text{pooled}} = \text{MaxPool}\left( \mathbf{F}_{\text{out}},\ 2 \times 2 \right).
\]

As shown in Figure \ref{fig:multiview_module}, the spatial view within the multiview attention mechanism is particularly essential for enhancing spatial feature consistency. By focusing on specific regions within the feature maps, the spatial view ensures robust encoding of intricate smoke shapes, such as those formed by narrow plumes or widespread quarry blast emissions. Similarly, the height-channel and width-channel views capture directional patterns and align encoded features with global contexts, enabling robust feature extraction for both narrow, tall plumes and wide, elongated patterns.

This strategic combination of multiview linear attention ensures that the encoder stages of SmokeNet (4-6) are adept at refining the variability and complexity of smoke patterns, delivering feature representations optimized for segmentation in the decoder.

\subsection{Decoder}

\paragraph{Decoder with Skip Connections}
The decoder stages in SmokeNet (Stages 4-6) progressively reconstruct the spatial resolution of the feature maps using transposed convolutions. To enhance segmentation precision, skip connections are employed to transfer enriched features from the encoder to the corresponding decoder stages. These skip connections combine the encoder output at the current stage with the upsampled output of the lower-stage skip connection using element-wise addition, integrating features from multiple levels to ensure dimensional alignment and retain critical details necessary for accurate smoke segmentation.

Let the output tensor from the encoder at stage $k$ be denoted as 
$\mathbf{F}_{\text{encoder}}^k \in \mathbb{R}^{N \times C_k \times H_k \times W_k}$, where $C_k$, $H_k$, and $W_k$ represent the number of channels, height, and width, respectively, at stage $k$. The output of the lower-stage skip connection is denoted as $\mathbf{F}_{\text{skip}}^{k+1} \in \mathbb{R}^{N \times C_{k+1} \times H_{k+1} \times W_{k+1}}$. The skip connection input at stage $k$ is computed by adding the encoder output with the upsampled lower-stage skip connection as follows:

\[
\mathbf{F}_{\text{skip}}^k = \mathbf{F}_{\text{encoder}}^k + \operatorname{Up}\left(\mathbf{F}_{\text{skip}}^{k+1}\right) \in \mathbb{R}^{N \times C_k \times H_k \times W_k},
\]
where $\operatorname{Up}(\cdot)$ represents an upsampling operation (e.g., transposed convolution) that aligns the spatial resolution of the lower-stage skip connection with the current stage.

\paragraph{Decoder Stage Operations}
At each decoder stage, the input $\mathbf{F}_{\text{decoder}}^k$ is formed by adding the output of the skip connection with the upsampled output of the lower decoder stage. This is mathematically expressed as:
\[
\mathbf{F}_{\text{decoder}}^k = \mathbf{F}_{\text{skip}}^k + \text{Up}\left(\mathbf{F}_{\text{decoder}}^{k+1}\right) \in \mathbb{R}^{N \times C_k \times H_k \times W_k},
\]
where $\mathbf{F}_{\text{decoder}}^{k+1} \in \mathbb{R}^{N \times C_{k}' \times H_{k+1} \times W_{k+1}}$ is the output from the lower decoder stage, and $\text{Up}(\cdot)$ ensures spatial alignment.

Once the decoder input is established, it undergoes a series of transposed convolutions and linear operations to reconstruct the spatial resolution while reducing the channel dimensions:
\[
\mathbf{F}_{\text{output}}^k = \text{TransposedConv}_{3 \times 3}\left(\mathbf{F}_{\text{decoder}}^k\right) \in \mathbb{R}^{N \times C_k' \times H_k \times W_k},
\]
where $C_k'$ represents the reduced channel dimension at stage $k$. This transposed convolution operation progressively upsamples the feature maps and diminishes the number of channels, effectively restoring spatial resolution while preserving essential details for accurate segmentation.

In the final stage of the decoder, a single-channel segmentation mask is produced through a transposed convolution followed by a sigmoid activation function:

\begin{align*}
\mathbf{F}_{\text{segmentation}} = \sigma\Bigg(&\text{TransposedConv}_{3 \times 3} \Big( \mathbf{F}_{\text{output}}^1 \Big) \Bigg) \\
&\in \mathbb{R}^{N \times 1 \times H \times W}.
\end{align*}

where $\sigma$ denotes the sigmoid activation function, ensuring that the output values are scaled between 0 and 1, suitable for binary segmentation tasks.

This structured use of skip connections and decoder operations ensures the seamless integration of multi-level features, enabling accurate smoke segmentation with fine spatial detail.

\subsection{Loss Function}

For training, we use a combined loss function that includes a binary cross-entropy (BCE) component and a Dice loss component, formulated to optimize segmentation accuracy by balancing region overlap and boundary alignment:

\begin{align*}
    \mathcal{L}_{\text{combined}} = \alpha \cdot \text{BCE}(y, \hat{y}) + \beta \cdot \text{Dice}(y, \hat{y})
\end{align*}

where \( y \) is the ground truth, \( \hat{y} \) is the predicted mask, and \(\alpha\) and \(\beta\) are weights assigned to each loss component to balance their contributions.

We further implement a cosine annealing learning rate schedule to modulate the learning rate during training, aiming to facilitate smoother convergence. The learning rate \( \eta_t \) at epoch \( t \) is given by:

\begin{align*}
    \eta_t = \eta_{\text{min}} + \frac{1}{2} (\eta_{\text{max}} - \eta_{\text{min}}) \left(1 + \cos\left(\frac{t}{T} \pi \right)\right)
\end{align*}

where \( \eta_{\text{min}} \) and \( \eta_{\text{max}} \) are the minimum and maximum learning rates, respectively, and \( T \) is the total number of epochs. For our model, we set \( \eta_{\text{max}} = 0.001 \), \( \eta_{\text{min}} = 1 \times 10^{-6} \), and \( T = 100 \) epochs.

With the loss function, we incorporate layer-wise loss functions that combine the overall loss at different layers. Since the layer loss is the combined loss at different layers, we assigned different weights to each layer's loss to balance their contributions. The layer-wise loss is defined as:

\begin{align*}
    \mathcal{L}_{\text{layer}} = \sum_{i=1}^{N} \gamma_i \cdot \mathcal{L}_{\text{combined}, i}
\end{align*}

where \( \mathcal{L}_{\text{combined}, i} \) represents the combined loss at layer \( i \), and \( \gamma_i \) is the weight assigned to layer \( i \). For the layer-wise loss, different weights were assigned to each layer to balance their contributions, with the values assigned as follows: Stage 2 to 6 has decreasing weights of \( \gamma_1 = 0.5 \), \( \gamma_2 = 0.4 \), \( \gamma_3 = 0.3 \), \( \gamma_4 = 0.2 \), \( \gamma_5 = 0.1 \).

\begin{table*}[ht]
\centering
\caption{Module evaluation results of model modules across 4 test datasets with mean and standard deviation.\textsuperscript{*}}
\label{table:module_comprehensive}
\small
\setlength{\tabcolsep}{4pt} 
\begin{tabular}{lcccccccccc}
\toprule
\textbf{Model} & \textbf{Conv} & \textbf{Loss} & \textbf{Attention} & \multicolumn{4}{c}{\textbf{mIoU (\%)}} & \textbf{\#Params (K) $\downarrow$} & \textbf{MFLOPs $\downarrow$} & \textbf{FPS $\uparrow$} \\
\cmidrule(lr){5-8}
 & & & & \textbf{Smoke100k} & \textbf{DS01} & \textbf{Fire Smoke} & \textbf{Quarry Smoke} & & & \\
\midrule
M1 & N & F  & NA & 72.40$\pm$0.08 & 70.83$\pm$0.06 & 70.45$\pm$0.11 & 69.12$\pm$0.05 & \multirow{2}{*}{417.63} & \multirow{2}{*}{236.02} & \multirow{2}{*}{54.25} \\
M2 & N & L  & NA & 70.75$\pm$0.09 & 67.45$\pm$0.06 & 66.16$\pm$0.08 & 63.52$\pm$0.07 & & & \\
\midrule
M3 & N & F  & MA  & 73.81$\pm$0.07 & 71.53$\pm$0.12 & 69.62$\pm$0.04 & 67.74$\pm$0.10 & \multirow{2}{*}{710.82} & \multirow{2}{*}{120.18} & \multirow{2}{*}{56.03} \\
M4 & N & L  & MA  & 74.10$\pm$0.07 & 73.14$\pm$0.06 & 72.24$\pm$0.05 & 71.67$\pm$0.08 & & & \\
\midrule
M5 & M & F  & NA & 72.19$\pm$0.10 & 69.78$\pm$0.05 & 67.22$\pm$0.07 & 63.71$\pm$0.06 & \multirow{2}{*}{\textbf{230.63}} & \multirow{2}{*}{84.72} & \multirow{2}{*}{\textbf{128.65}} \\
M6 & M & L  & NA & 72.24$\pm$0.04 & 71.41$\pm$0.05 & 68.95$\pm$0.08 & 66.67$\pm$0.07 & & & \\
\midrule
M7 & M & F  & MA  & 75.63$\pm$0.05 & 73.83$\pm$0.09 & 71.22$\pm$0.06 & 70.34$\pm$0.07 & \multirow{2}{*}{344.64} & \multirow{2}{*}{\textbf{74.51}} & \multirow{2}{*}{77.05} \\
M8 & M & L  & MA  & \textbf{76.45$\pm$0.10} & \textbf{74.43$\pm$0.04} & \textbf{73.43$\pm$0.03} & \textbf{72.74$\pm$0.06} & & & \\
\bottomrule
\end{tabular}

\vspace{2mm}

\scriptsize{\textsuperscript{*} \textbf{Model}: M1-M8 represent different combinations of components; \textbf{Conv}: N = Normal Convolution, M = Multiscale Convolution; \textbf{Loss}: F = Final Output Loss Only, L = Layer-Specific Losses; \textbf{Attention}: MA = Multiview Linear Attention, NA = Normal Linear Attention.}
\end{table*}

\begin{table*}[htbp]  
\centering
\caption{Detailed comparison of various methods for semantic segmentation with mean and standard deviation.}
\label{tab:comparison_methods}
\small  
\setlength{\tabcolsep}{4pt} 
\begin{tabular*}{\textwidth}{@{\extracolsep{\fill}} 
    p{3.5cm}  
    c c c c  
    c       
    c       
    c       
}

\toprule
\textbf{Methods} & \multicolumn{4}{c}{\textbf{mIoU (\%)}} & \textbf{\#Params (M) $\downarrow$} & \textbf{GFLOPs $\downarrow$} & \textbf{FPS $\uparrow$} \\
\cmidrule(lr){2-5}
 & \textbf{Smoke100k} & \textbf{DS01} & \textbf{Fire Smoke} & \textbf{Quarry Smoke} & & & \\
\midrule
UNet (2015) & 66.13$\pm$0.10 & 61.32$\pm$0.08 & 60.14$\pm$0.06 & 57.18$\pm$0.05 & 28.24 & 35.24 & 75.58 \\
UNet++ (2018) & 69.12$\pm$0.09 & 64.65$\pm$0.04 & 61.77$\pm$0.10 & 58.44$\pm$0.06 & 9.16  & 10.72 & 91.25 \\
AttentionUNet (2018) & 69.68$\pm$0.05 & 66.59$\pm$0.12 & 64.15$\pm$0.07 & 59.64$\pm$0.09 & 31.55 & 37.83 & 46.48 \\
UNeXt-S (2022) & 72.25$\pm$0.11 & 71.62$\pm$0.07 & 69.59$\pm$0.12 & 64.54$\pm$0.04 & 0.77 & 0.08 & \textbf{202.06} \\
MobileViTv2 (2022) & 71.73$\pm$0.10 & 71.54$\pm$0.07 & 70.23$\pm$0.04 & 69.12$\pm$0.11 & 2.30 & 0.09 & 98.84 \\
MALUNet (2022) & 71.81$\pm$0.05 & 70.16$\pm$0.10 & 69.42$\pm$0.04 & 67.64$\pm$0.07 & \textbf{0.17} & 0.09 & 87.72 \\
ERFNet (2017) & 71.84$\pm$0.09 & 71.38$\pm$0.10 & 66.59$\pm$0.06 & 66.24$\pm$0.07 & 2.06 & 3.32 & 61.22 \\
LEDNet (2019) & 70.76$\pm$0.10 & 71.63$\pm$0.07 & 70.13$\pm$0.08 & 67.74$\pm$0.11 & 0.91 & 1.41 & 60.19 \\
DFANet (2019) & 66.91$\pm$0.05 & 63.87$\pm$0.10 & 62.76$\pm$0.07 & 70.21$\pm$0.09 & 2.18 & 0.44 & 31.05 \\
CGNet (2020) & 75.64$\pm$0.07 & 73.76$\pm$0.11 & 72.04$\pm$0.10 & 71.91$\pm$0.08 & 0.49 & 0.86 & 53.53 \\
DSS (2019) & 73.25$\pm$0.05 & 72.17$\pm$0.04 & 69.78$\pm$0.12 & 69.81$\pm$0.07 & 30.20 & 184.90 & 32.56 \\
Frizzi (2021) & 73.44$\pm$0.06 & 71.67$\pm$0.09 & 70.51$\pm$0.07 & 70.40$\pm$0.11 & 20.17 & 27.90 & 60.32 \\
Yuan (2023) & 75.57$\pm$0.07 & \textbf{74.84$\pm$0.06} & 71.94$\pm$0.10 & 70.92$\pm$0.08 & 0.88 & 1.15 & 68.81 \\ 
SmokeNet &\textbf{76.45$\pm$0.10} & 74.43$\pm$0.04 & \textbf{73.43$\pm$0.03} & \textbf{72.74$\pm$0.06} & 0.34  & \textbf{0.07}  & 77.05 \\
\bottomrule
\end{tabular*}
\end{table*}

\begin{figure*}[ht]
    \centering
    \includegraphics[width=1.0\linewidth]{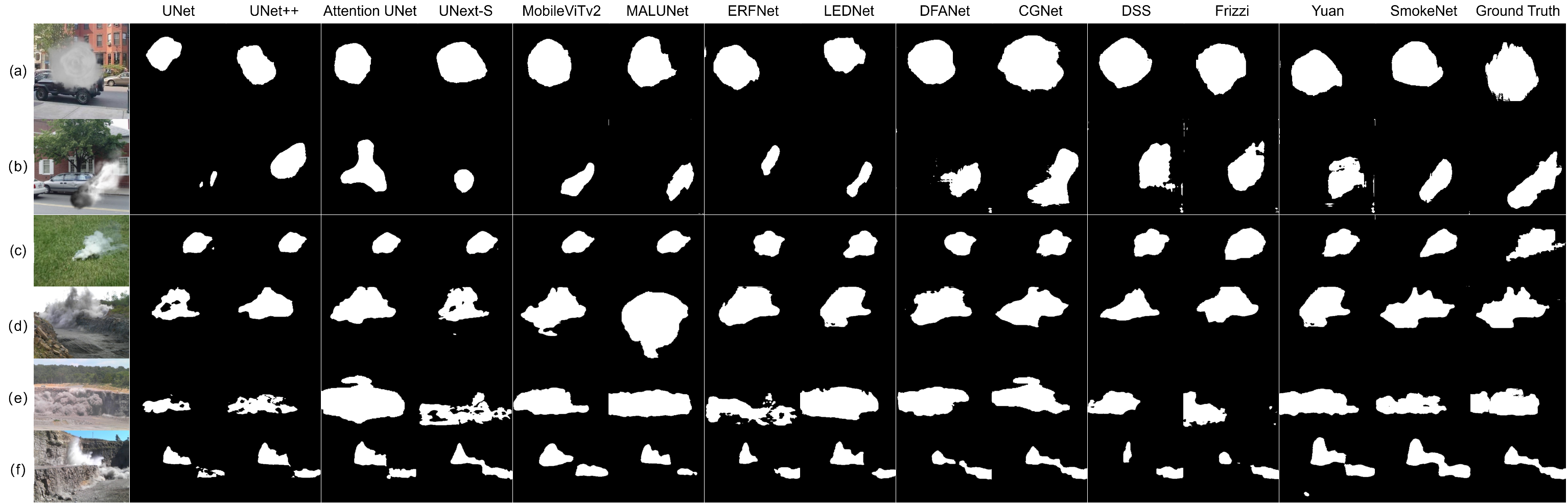}
    \caption{Segmentation results of SmokeNet and comparison models on sample images from four test datasets.}
    \label{fig:sample_images}
\end{figure*}

\section{Experiments}
Our experimental section was designed to answer the following research questions:

\begin{itemize}
    \item[\textbf{RQ1:}] How do SmokeNet’s multiscale convolutions, multiview linear attention, and layer-wise losses enhance its ability to capture complex smoke features?
    
    \item[\textbf{RQ2:}] How does SmokeNet's segmentation performance (mIoU) compare to established models across various datasets, including Quarry Smoke?
    
    \item[\textbf{RQ3:}] How does SmokeNet balance efficiency and accuracy compared to heavyweight and lightweight models in terms of parameters, GFLOPs, and FPS?
\end{itemize}

\subsection{Experimental Setup}
\paragraph{Deep Learning Architecture} 
SmokeNet was implemented using the PyTorch framework and trained on a single NVIDIA Tesla P40 24GB GPU. The model consists of six encoder and decoder layers with filter sizes \(\in [4, 8, 16, 32, 64, 128]\). Training was conducted using the AdamW optimizer with a learning rate of 0.001, a weight decay of \(1 \times 10^{-5}\), and a cosine annealing learning rate schedule (\(\eta_{\text{min}} = 0.00001\), \(T_{\text{max}} = 50\) iterations). The model was trained for 100 epochs with a batch size of 8, using a combined loss of cross-entropy and dice loss with layer-wise losses as the loss function. 
\paragraph{Dataset} Four datasets selected to encompass both synthetic and real-world smoke variations, testing its robustness and adaptability across diverse conditions:

\begin{itemize}
    \item \textbf{Smoke100k-M} \cite{cheng2019smoke}: A synthetic dataset comprising 25,000 training images and 15,000 test images, as illustrated in Figure~\ref{fig:sample_images} (a). It features centrally located smoke plumes of fixed size, providing a baseline for accuracy assessment in controlled conditions.

    \item \textbf{DS01} \cite{yuan2023lightweight}: This dataset comprises 70,632 training images and 1,000 test images, featuring various smoke sizes and locations, as illustrated in Figure~\ref{fig:sample_images} (b). It challenges the model’s ability to adapt to different spatial configurations, enhancing its generalization capabilities.

    \item \textbf{Fire Smoke} \cite{kaabi2020efficient}: A real-world dataset with 3,826 images, including 3,060 training images and 766 test images, as illustrated in Figure~\ref{fig:sample_images} (c). It captures both outdoor wildfire smoke and indoor smoke scenarios, providing realistic environments where smoke detection is critical for early fire warning and safety monitoring.

    \item \textbf{Quarry Smoke}: An industrial dataset comprising 3,703 images, including 2,962 training images and 741 test images, as illustrated in Figures~\ref{fig:sample_images} (d), (e), and (f). It represents dense, irregular smoke plumes mixed with dust and debris from quarry blasts, testing the model’s ability to segment smoke in dynamic and high-variability environments.
\end{itemize}
\paragraph{Data Augmentation} 
To enhance SmokeNet's robustness and generalization capabilities, we employed a comprehensive data augmentation pipeline that includes both basic and enhanced augmentation techniques.

Basic augmentations, including random horizontal and vertical flips, rotations, and brightness adjustments, were applied uniformly across all datasets to introduce general variability and prevent overfitting. In addition to these common techniques, we implemented enhanced augmentations—synthetic fog and motion blur—to address domain-specific challenges in our datasets.

Synthetic fog better simulates real-world scenarios, such as blast events occurring under overcast skies, rainy showers, and high humidity in mountainous regions. Motion blur is particularly relevant in real-world situations like quarry sites, where hand-held cameras or ground cameras mounted on tripods may experience distortions due to ground vibrations caused by explosive energy.

These enhanced augmentations were crucial in enabling SmokeNet to generalize effectively across both synthetic and real-world conditions, leading to significant improvements in smoke segmentation performance.


\paragraph{Performance Metrics}

SmokeNet's performance was evaluated using four metrics:

\begin{itemize}
    \item \textbf{Mean Intersection over Union (mIoU)}: Assesses segmentation accuracy by quantifying the overlap between predicted and ground-truth masks.
    \item \textbf{Parameter Count}: Indicates model scalability and resource usage. Reported in millions (M) or thousands (K), with units specified in each table.
    \item \textbf{Floating Point Operations (FLOPs)}: Measures computational complexity. Reported in gigaflops (GFLOPs) or megaflops (MFLOPs), as indicated in the tables.
    \item \textbf{Frames per Second (FPS)}: Reflects inference speed, critical for computationally constrained applications in dynamic environments like quarry blast monitoring.
\end{itemize}


\subsection{Results and Discussion}
\paragraph{Results} 
As shown in Table~\ref{table:module_comprehensive}, we evaluated different configurations of our model, investigating the contributions of multiscale convolution, multiview attention, and layer-specific loss functions. Each configuration was trained and evaluated on fixed, predefined training and testing splits of the datasets, with five independent runs conducted for each. The reported results include the mean and standard deviation, demonstrating the stability and effectiveness of these design choices.

In Table~\ref{tab:comparison_methods}, we compared the best-performing configuration of our model with several state-of-the-art segmentation methods. For each method, the same experimental protocol was applied, ensuring consistency in training and evaluation across the standardized dataset splits and five repetitions. This provides a rigorous and fair comparison, reflecting both segmentation accuracy and computational efficiency.

Figure~\ref{fig:sample_images} illustrates qualitative segmentation results across diverse datasets, including synthetic smoke with uniform, circular patterns; real-world fire smoke with irregular and amorphous structures; and quarry blast smoke characterized by dense, complex plumes. These visualizations offer insights into the strengths and limitations of our model in handling varied smoke characteristics, showcasing its adaptability across synthetic and real-world scenarios.

\paragraph{Impact of Architectural Innovations}

An module evaluation study was conducted to evaluate the contribution of SmokeNet’s multiscale convolutional layers and multiview linear attention mechanisms to its segmentation performance and efficiency across four datasets: Smoke100k, DS01, Fire Smoke, and Quarry Smoke. The study compared various model configurations (M1 through M8), as detailed in Table~\ref{table:module_comprehensive}. Models incorporating multiscale convolutions (M5-M8) consistently outperformed those with normal convolutions (M1-M4), with configuration M8 achieving an mIoU of 76.45\% on Smoke100k compared to 72.40\% for M1. Additionally, the integration of the multiview linear attention mechanism (present in M3, M4, M7, M8) further enhanced performance, with M8 attaining the highest mIoU scores across all datasets, including 72.74\% on Quarry Smoke. The inclusion of layer-specific loss functions also contributed to improved accuracy, as seen in configurations M2, M4, M6, and M8. Overall, the optimal configuration (M8), which combines multiscale convolutions, multiview linear attention, and layer-specific losses, demonstrated the most significant performance gains while maintaining low computational demands (344.64K parameters and 74.51 MFLOPs) and high inference speed (77.05 FPS). This validates the effectiveness of SmokeNet’s architectural innovations in capturing complex smoke features in dynamic environments \textbf{(RQ1 answered)}. 


\paragraph{Segmentation Performance Comparison}
As shown in Table~\ref{tab:comparison_methods}, SmokeNet demonstrates consistently high performance across multiple datasets. On the Smoke100k dataset, it achieves the highest mIoU of 76.45\%, outperforming CGNet (75.64\%) and Yuan (75.57\%), while significantly surpassing other models such as UNeXt-S (72.25\%) and MobileViTv2 (71.73\%). On the Fire Smoke dataset, SmokeNet attains an mIoU of 73.43\%, exceeding CGNet's 72.04\% and MobileViTv2's 70.23\%, and outperforming other models like Frizzi (70.51\%) and Yuan (71.94\%). Similarly, in the Quarry Smoke dataset, SmokeNet achieves an mIoU of 72.74\%, surpassing CGNet (71.91\%), Yuan (70.92\%), and Frizzi (70.40\%). However, on the DS01 dataset, SmokeNet achieves an mIoU of 74.43\%, slightly below Yuan's 74.84\%. Despite this, it maintains competitiveness by outperforming models such as CGNet (73.76\%), Frizzi (71.67\%), and UNeXt-S (71.62\%). Overall, SmokeNet demonstrates robustness and generalizability, achieving consistent improvements over lightweight models like MALUNet (70.16\%) and LEDNet (71.63\%), while also outperforming computationally heavier models such as AttentionUNet (66.59\%) and DSS (72.17\%) across most datasets.

In Figure~\ref{fig:sample_images}, SmokeNet exhibits its ability to accurately delineate complex smoke boundaries, capture fine-grained details, and maintain consistent segmentation across various smoke scenarios, including challenging quarry smoke environments. Traditional models like UNet and UNet++ produce simplified masks that miss intricate contours and fragmented structures. Smoke segmentation-specific models such as DSS, Frizzi, and Yuan also generate less precise masks compared to SmokeNet. Lightweight models such as UNeXt-S and MobileViTv2 often overlook subtle edges and fine details. For the quarry smoke case application, which is commonly used for pollutant quantification inside the smoke plume, slightly thicker masks than the ground truth are acceptable compared to missing parts, which could potentially omit noxious chemicals of the smoke, as shown in Figure~\ref{fig:sample_images} (d), (e), and (f). In contrast, SmokeNet consistently generates segmentation masks closely aligned with the ground truth, effectively capturing irregular shapes, narrow projections, and diffused edges.

\begin{figure}[ht]
    \centering
    \includegraphics[width=1.0\linewidth]{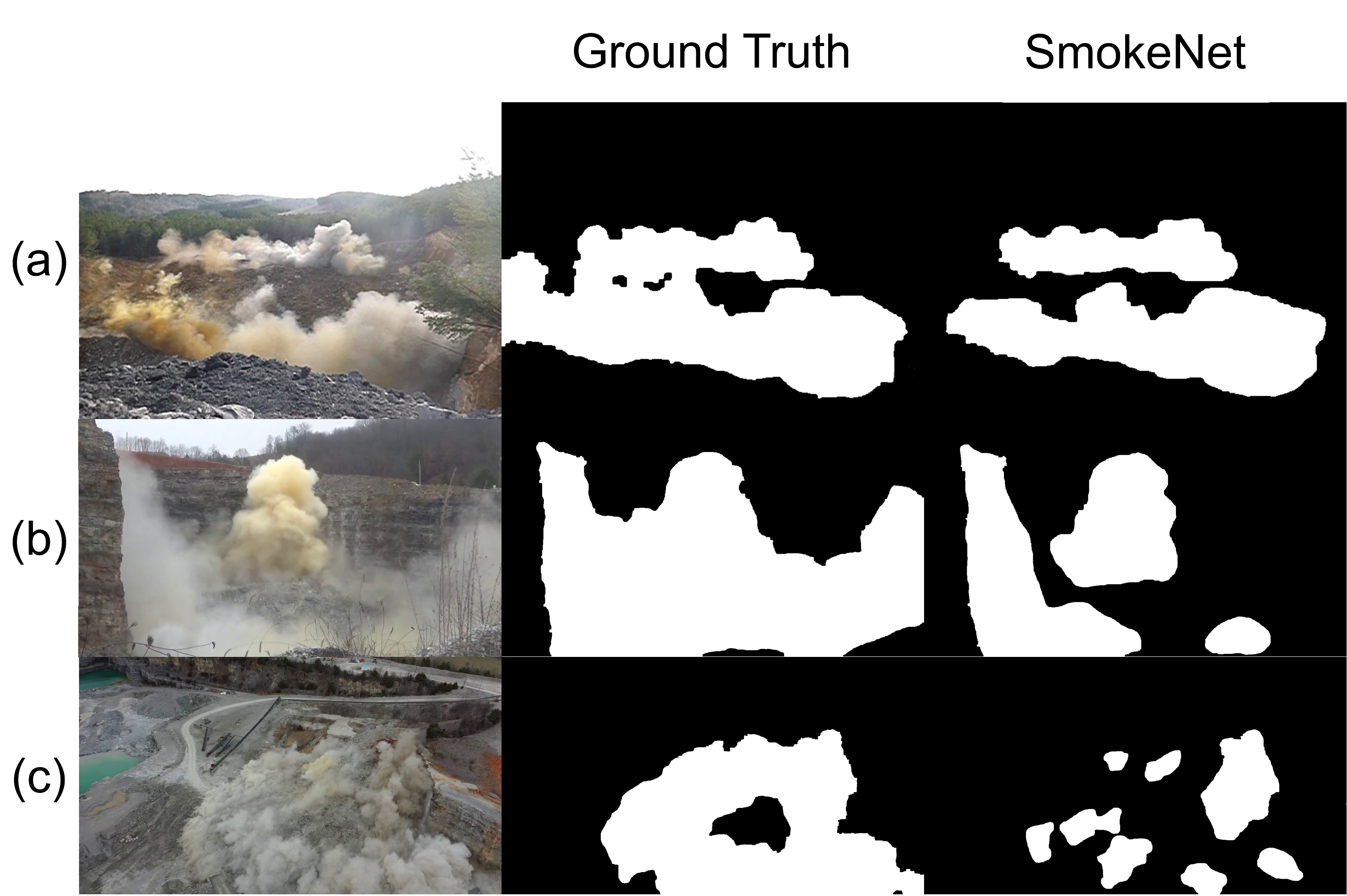}
    \caption{Segmentation results of SmokeNet in challenging quarry scenarios. 
    }
    \label{fig:failure_cases}
\end{figure}

Figure~\ref{fig:failure_cases} highlights the performance of SmokeNet across different challenging scenarios where SmokeNet does not effectively segment the entire smoke in these images. In Figure~\ref{fig:failure_cases}(a), the image contains two horizontal clusters of smoke, with most of them being white and gray under bright sunlight. However, for the orange part on the left side of the bottom smoke cluster, SmokeNet recognizes the dark orange portion of the smoke but misses the faint, translucent orange smoke, which is less observable, especially against the orange soil background, due to the gradual transitions between smoke and background. In Figure~\ref{fig:failure_cases}(b), the ground camera provides a closer view of the smoke plume at ground level. The complexity of the background textures and the irregularity of the smoke shapes make it difficult for SmokeNet to accurately delineate the two-part smoke composition: the yellowish smoke directly emanating from the collapsed rocks during the blast, and the white smoke surrounding it, which falls to the lower ground level first and then disperses around. Additionally, the presence of dry twigs in front of the smoke further confuses the model in recognizing the entire shape of the smoke, especially when combined with the faint smoke in the center and the rocks observed behind. In Figure~\ref{fig:failure_cases}(c), the smoke image is captured from a drone-mounted camera, which covers the overall view of the smoke plume spread across the entire quarry site. SmokeNet can recognize the smoke near the blast spot and the rock wall but cannot accurately recognize the dispersed smaller grouped smoke clusters due to the low visibility of sparse smoke regions and the intricate gray background surface textures.

Despite these challenges requiring further improvement, the overall quantitative and qualitative analysis demonstrates that SmokeNet performs better than established models in segmentation performance across various datasets. The higher mIoU scores and consistent qualitative results highlight SmokeNet's effectiveness in accurately segmenting smoke under diverse conditions \textbf{(RQ2 answered)}.

\paragraph{Model Efficiency Comparison}
Table~\ref{tab:comparison_methods} provides a comprehensive comparison of various semantic segmentation methods based on number of parameters (\#Params), computational complexity (GFLOPs), and inference speed (FPS), while also considering their segmentation performance (mIoU).

SmokeNet demonstrates exceptional efficiency with only 0.34M parameters and the lowest computational complexity of 0.07 GFLOPs, achieving an inference speed of 77.05 FPS. This makes SmokeNet one of the most lightweight and computationally efficient models among the compared methods. Additionally, SmokeNet maintains competitive mIoU scores, outperforming other models on Smoke100k, Fire Smoke, and Quarry Smoke datasets.

Traditional models like UNet and UNet++ have significantly higher parameter counts (28.24M and 9.16M, respectively) and GFLOPs (35.24 and 10.72), which result in heavy workloads for GPU memory and computing usage. AttentionUNet offers slightly improved mIoU scores but at the cost of increased parameters (31.55M) and GFLOPs (37.83 GFLOPs), leading to a slower inference speed of 46.48 FPS. Additionally, models such as ERFNet and DFANet have lower parameter counts compared to UNet and UNet++ (2.06M and 2.18M, respectively), but still maintain relatively higher GFLOPs (3.32 and 0.44 GFLOPs) and lower FPS (61.22 and 31.05), resulting in significant computational demands compared to most other models.

Advanced models such as UNeXt-S and MobileViTv2 achieve higher inference speeds of 202.06 FPS and 98.84 FPS with 0.77M and 2.30M parameters, and 0.08 GFLOPs and 0.09 GFLOPs, respectively. However, their mIoU scores are generally lower than those of SmokeNet.

Among parameter-efficient models, MALUNet is the most lightweight with only 0.17M parameters and 0.09 GFLOPs, while CGNet and LEDNet offer a balance between low parameter counts (0.49M and 0.91M) and reasonable GFLOPs (0.86 and 1.41). Nonetheless, SmokeNet outperforms these models in terms of computational efficiency and maintains competitive inference speeds.

Models specifically designed for smoke segmentation, such as DSS and Frizzi, require significantly higher GFLOPs (184.90 GFLOPs and 27.90 GFLOPs, respectively) and exhibit slower inference speeds (32.56 FPS and 60.32 FPS). While achieving respectable mIoU scores, their computational demands are considerably higher compared to SmokeNet. Yuan achieves competitive mIoU scores with 0.88M parameters and 1.15 GFLOPs, but SmokeNet generally offers a better efficiency-performance balance across most datasets, except for the DS01 dataset where Yuan achieves the highest mIoU. 

Overall, SmokeNet provides the best trade-off between computational demand and segmentation performance, making it highly suitable for real-time smoke segmentation tasks \textbf{(RQ3 answered)}.

\section{Conclusion}

This study introduces SmokeNet, an efficient and robust model for smoke plume segmentation across diverse scenarios, including quarry blast smoke. By integrating multiscale convolutions and multiview linear attention within a lightweight framework, SmokeNet effectively handles dynamic smoke plumes with varying opacity and shape. Experimental results demonstrate high segmentation accuracy on both synthetic and real-world datasets, such as campfire, wildfire, and quarry blast smoke. Its low parameter count, reduced computational demands, and high inference speed make it suitable for applications in environmental monitoring and industrial safety.

However, the failure case study highlighted limitations in identifying sparse smoke regions against complex backgrounds and managing irregular smoke shapes and low visibility areas. Addressing these challenges will involve enhancing feature extraction techniques, improving background differentiation, utilizing augmented and synthetic datasets for greater robustness, and optimizing SmokeNet's architecture for real-time processing. Exploring dynamic kernel shapes may also improve generalizability for irregular objects like smoke plumes. Overall, SmokeNet offers a balanced trade-off between performance and computational efficiency, making it a valuable tool for real-time smoke detection and monitoring in various applications.

\section{Acknowledgements}
This work was supported by Austin Powder. We gratefully acknowledge Austin Powder for providing the essential dataset that enabled the development and evaluation of our approach.

\bibliography{aaai25}

\end{document}